\documentclass[conference]{IEEEtran}
\IEEEoverridecommandlockouts
\usepackage{cite}
\usepackage{amsmath,amssymb,amsfonts}
\usepackage{graphicx}
\usepackage{textcomp}
\usepackage{xcolor}
\usepackage{xspace}
\usepackage{xurl}
\def\BibTeX{{\rm B\kern-.05em{\sc i\kern-.025em b}\kern-.08em
    T\kern-.1667em\lower.7ex\hbox{E}\kern-.125emX}}
\usepackage{mathtools}
\usepackage{multirow}
\usepackage{adjustbox}
\usepackage{tikz}
\usetikzlibrary{positioning}
\usepackage{hyperref}
\usepackage{caption}
\usepackage{subcaption}
\usepackage[french,boxed,vlined,linesnumbered,inoutnumbered,rightnl,algo2e]{algorithm2e}
\usepackage{algorithm}
\usepackage{mwe}
\usepackage{stfloats}
\usepackage{pifont}
\usepackage[table]{xcolor}

\begin{document}
\title{Enabling Vibration-Based Gesture Recognition on Everyday Furniture via Energy-Efficient FPGA Implementation of 1D Convolutional Networks}
\author{
    \IEEEauthorblockN{
    Koki Shibata\IEEEauthorrefmark{1}\textsuperscript{1}, 
    Tianheng Ling\IEEEauthorrefmark{1}\textsuperscript{2,3}, 
    Chao Qian\IEEEauthorrefmark{1}\textsuperscript{2,3},  \\
    Tomokazu Matsui\textsuperscript{1,4},
    Hirohiko Suwa\textsuperscript{1,4},
    Keiichi Yasumoto\textsuperscript{1,4} and
    Gregor Schiele\textsuperscript{2,3}
    }
    \IEEEauthorblockA{
    \textsuperscript{1} Ubiquitous Computing Systems Laboratory, Nara Institute of Science and Technology, Nara, Japan\\
    \textsuperscript{2} Intelligent Embedded Systems Laboratory, University of Duisburg-Essen, Duisburg, Germany\\
    \textsuperscript{3} PALUNO, The Ruhr Institute for Software Technology, Essen, Germany\\
    \textsuperscript{4} RIKEN, Center for Advanced Intelligence Project, Tokyo, Japan\\
    }
    \IEEEauthorblockA{\IEEEauthorrefmark{1} These authors contributed equally to this work.}
}
\maketitle
\begin{abstract}
The growing demand for smart home interfaces has increased interest in non-intrusive sensing methods like vibration-based gesture recognition. While prior studies demonstrated feasibility, they often rely on complex preprocessing and large Neural Networks (NNs) requiring costly high-performance hardware, resulting in high energy usage and limited real-world deployability.
This study proposes an energy-efficient solution deploying compact NNs on low-power Field-Programmable Gate Arrays (FPGAs) to enable real-time gesture recognition with competitive accuracy. We adopt a series of optimizations:
(1) We replace complex spectral preprocessing with raw
waveform input, eliminating complex on-board preprocessing while reducing input size by 21$\times$ without sacrificing accuracy.
(2) We design two lightweight architectures (1D-CNN and 1D-SepCNN) tailored for embedded FPGAs, reducing parameters from 369 million to as few as 216 while maintaining comparable accuracy.
(3) With integer-only quantization and automated RTL generation, we achieve seamless FPGA deployment. A ping-pong buffering mechanism in 1D-SepCNN further improves deployability under tight memory constraints.
(4) We extend a hardware-aware search framework to support constraint-driven model configuration selection, considering accuracy, deployability, latency, and energy consumption.
Evaluated on two swipe-direction datasets with multiple users and ordinary tables, our approach achieves low-latency, energy-efficient inference on the AMD Spartan-7 XC7S25 FPGA. Under the PS data splitting setting, the selected 6-bit 1D-CNN reaches 0.970 average accuracy across users with 9.22~ms latency. The chosen 8-bit 1D-SepCNN further reduces latency to 6.83~ms (over 53$\times$ CPU speedup) with slightly lower accuracy (0.949). Both consume under 1.2~mJ per inference, demonstrating suitability for long-term edge operation.
All source code is available in the accompanying GitHub repository\footnote{\url{https://github.com/tianheng-ling/Smatable}}.
\end{abstract}


\section{Introduction}
\label{sec:introduction} 

The growing demand for smart home technologies has accelerated the development of interactive household interfaces, including vision-based, voice-controlled, and touch-enabled systems~\cite{shi2022progress}. These systems enhance convenience by enabling intuitive control of lighting, climate, and appliances. However, conventional interaction methods rely on cameras, microphones, or capacitive sensors, each with inherent limitations. Camera-based approaches provide precise gesture and motion recognition but are sensitive to occlusion and lighting conditions, while also raising privacy concerns~\cite{camera1,camera2}. Microphone-based systems enable hands-free use but are susceptible to ambient noise and pose privacy risks due to continuous audio recording~\cite{microphone1,microphone2}. Capacitive sensors, widely used in touch panels, do not work reliably on thick or metallic surfaces, limiting their integration into furniture~\cite{capacitance1,capacitance2}.

\begin{figure}[!htb]
\vspace{-15pt}
    \centering
    \includegraphics[width=.75\columnwidth]{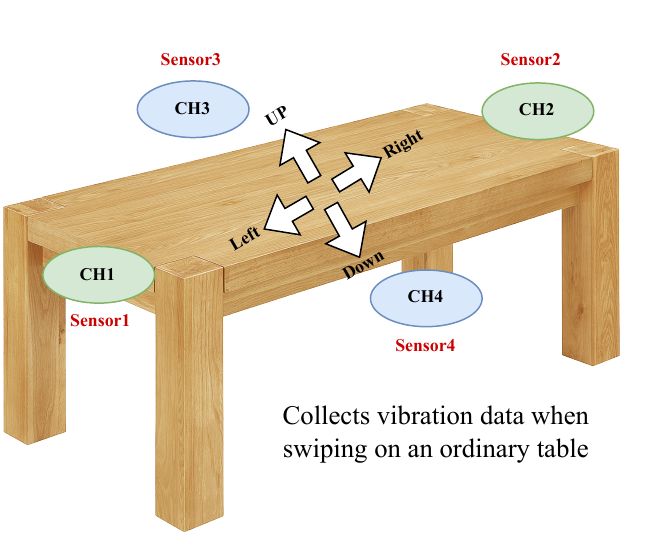}
    \caption{Illustration of Vibration-Sensing Setup for Swipe Recognition on an Ordinary Table Mounted by Four Sensors}
    \label{fig:tables} 
\end{figure}

To address these issues, Yoshida et al.\cite{yoshida2023smatable1} proposed \emph{Smatable} (see Figure \ref{fig:tables}), a vibration-based sensing system that transforms ordinary tables into interactive surfaces. By attaching vibration sensors to the underside of tables, \emph{Smatable} enables swipe gesture recognition without cameras or microphones, thereby improving privacy and material compatibility. Their follow-up work \cite{yoshida2023smatable2} improved accuracy by tuning \emph{Short-Time Fourier Transform} (STFT) parameters for feature extraction and optimizing a \emph{2D Convolutional Neural Network} (2D-CNN). However, the enhanced model requires 1.38 GB of memory and relies on a costly high-performance CPU with high energy consumption. Their approach incurs a latency of 15~ms for STFT preprocessing and 350~ms for inference on an Intel Core i7-12700H, resulting in a total delay of 365~ms, well above the sub-100~ms threshold required for fluid interaction~\cite{ng2012designing}.

These constraints motivate our research question: \textbf{Can accurate, low-latency, and energy-efficient gesture recognition be achieved on IoT-grade hardware for seamless furniture integration?} To this end, we focus on \emph{Field-Programmable Gate Arrays} (FPGAs), which offer a compelling balance of latency and energy efficiency for embedded applications. However, deploying large-scale \emph{Neural Networks} (NNs) on such resource-constrained platforms remains challenging, requiring careful software-hardware co-design. Through targeted optimizations in sensing, modeling, quantization, and deployment, we make the following contributions:

\begin{itemize}
    \item We replace complex spectral preprocessing with raw waveform input, eliminating the need for STFT. This reduces the input size by 21$\times$ without sacrificing accuracy.
    
    \item We propose two compact 1D convolutional architectures tailored for embedded FPGAs: a standard 1D \emph{Convolutional Network} (1D-CNN) and a \emph{Depthwise-Separable 1D Convolutional Network} (1D-SepCNN). Compared to the 2D-CNN baseline~\cite{yoshida2023smatable2} with over 369 million parameters, our models reduce the parameter count to just 216 while achieving comparable accuracy on the same datasets.
    
    \item By applying integer-only quantization and customized RTL templates, we enable seamless model deployment on embedded FPGAs. To further address memory bottlenecks, we integrate a ping-pong buffering mechanism into the 1D-SepCNN implementation, significantly enhancing its deployability on resource-constrained platforms.

    \item We extend a bi-objective, hardware-aware Optuna-based search framework to support constraint-driven model selection, enabling automatic model configuration that balance deployment requirements across accuracy, deployability, latency, and energy usage.
    
    \item We validate our solution on two swipe-direction datasets involving multiple users and tables. Under the PS data split setting, the selected 6-bit 1D-CNN achieves 0.970 accuracy with a 9.22 ms latency. The 8-bit 1D-SepCNN further reduces latency to 6.83 ms (over 53$\times$ speedup over CPU-based inference\cite{yoshida2023smatable2}) while maintaining comparable accuracy (0.949). Both models consume less than 1.2 mJ per inference, demonstrating strong suitability for energy-constrained edge deployment.
    
\end{itemize}

The remainder of this paper is organized as follows: 
Section~\ref{sec:related_work} reviews related literature. 
Section~\ref{sec:proposed_method} describes our deployment-oriented optimization techniques.
Section~\ref{sec:results_eval} describes the experimental setup and evaluates the results.
Section~\ref{sec:conclusion_future_work} concludes the study and outlines future directions.

\section{Related Work}
\label{sec:related_work}

This section reviews relevant research along three directions: vibration-based interaction techniques, efficient NNs for embedded inference, and quantization and hardware acceleration methods for resource-constrained deployment.

\subsection{Vibration-based Interaction}

Vibration-based sensing has emerged as a promising modality for enabling unobtrusive, material-agnostic, and privacy-preserving smart interfaces~\cite{huang2020vibration}. Early approaches typically relied on handcrafted features and traditional \emph{Machine Learning} (ML) classifiers. For instance, \emph{SurfaceVibe}~\cite{pan2017surfacevibe} estimated wave propagation delays to detect taps and swipes, but suffered from limited adaptability due to fixed signal processing heuristics. \emph{VibSense}~\cite{liu2017vibsense} extracted time- and frequency-domain features and used classifiers such as \emph{Support Vector Machines} and \emph{Random Forests}, while \emph{iWood}~\cite{wu2022iwood} embedded piezoelectric sensors into furniture and applied tree-based models for recognition. Although effective, these systems rely on manual feature design, limiting scalability and generalization.

To overcome these limitations, more recent work has turned to \emph{Deep Learning}. Yoshida et al.~\cite{yoshida2023smatable1,yoshida2023smatable2} demonstrated the feasibility of vibration-based gesture recognition using large 2D-CNNs applied to STFT-based spectrograms. While their system achieves high accuracy in controlled settings, it depends on computationally intensive preprocessing and high-capacity models, resulting in substantial latency and energy overhead. These drawbacks pose barriers to practical deployment on embedded devices. Our work addresses these challenges by rethinking the entire sensing-to-inference pipeline from a deployment perspective, aiming to achieve accurate, real-time recognition under stringent hardware constraints.

\subsection{Compact Neural Architectures for Embedded Inference}

To support deployment on constrained devices, NNs must strike a balance between recognition accuracy and computational efficiency.
One often overlooked factor is the dimensionality of the input data. Larger inputs increase memory usage and amplify the computational burden in early network layers~\cite{menghani2023efficient}. In vibration-based sensing, high-resolution inputs such as spectrograms contribute significantly to system overhead. Inspired by trends in speech and audio processing~\cite{nossier2020comparative}, we consider direct processing of downsampled raw waveforms, which can preserve essential temporal patterns while reducing input size and enabling real-time inference.

Architectural efficiency is equally critical. While prior work~\cite{yoshida2023smatable2} adopts 2D-CNNs, several studies suggest that 1D-CNNs are better suited for sequential signals like vibration data~\cite{kiranyaz20191, shahid2022performance}, offering a favorable trade-off between expressiveness and resource efficiency. Further improvements can be achieved through depthwise-separable convolutions, which have proven effective in embedded contexts~\cite{mo2021fpga}. These insights motivate our choice of both standard and separable 1D-CNN architectures tailored to the deployment constraints.

\subsection{Quantization and FPGA Deployment}

Beyond compact architectures, embedded deployment requires system-level optimizations that account for hardware constraints.
Quantization is a widely adopted technique to reduce model size and computation by converting floating-point operations into low-bit integer arithmetic (e.g., INT8)~\cite{Jacob2017, Krishnamoorthi2018}. Integer-only inference is particularly advantageous on FPGAs, where custom datapaths can exploit reduced bitwidths to improve energy efficiency and throughput~\cite{ling2025deployment}.

Moreover, hardware acceleration using FPGAs has shown great promise for achieving real-time, energy-efficient inference. For example, Rashid et al.~\cite{Rashid2024} proposed \emph{TinyM2Net-V2}, which leverages sparsity and hybrid quantization to accelerate inference on an \emph{Artix-7} FPGA. Liao et al.~\cite{Liao2024} introduced \emph{BearingPGA-Net}, which uses parallel processing to speed up CNN-based fault detection on a \emph{Kintex-7} FPGA. While these works highlight the benefits of FPGA acceleration, they typically focus on high-performance FPGAs with ample computational resources.
In contrast, fewer studies have investigated deployment on low-power embedded FPGAs such as the \emph{AMD Spartan-7}, where resource budgets are much more constrained. These efforts underscore the need for end-to-end co-design strategies that align model structure, quantization, and hardware capabilities. Building on these insights, our work targets low-power FPGAs and explores quantization-aware model architecture and optimization techniques to enable efficient real-time gesture recognition using vibration data.

\section{Deployment-Oriented Optimizations}
\label{sec:proposed_method}

This section introduces a series of deployment-oriented optimizations for real-time swipe gesture recognition on embedded FPGAs. We first replace high-cost spectral features with downsampled raw waveforms to minimize preprocessing overhead. Compact 1D-CNN and 1D-SepCNN architectures are then designed to meet tight hardware constraints. To support efficient deployment, we apply \emph{Quantization-Aware Training} (QAT) and generate hardware accelerators using template-based RTL design. Finally, we incorporate constraint-aware, bi-objective optimization into an end-to-end deployment framework, enabling Optuna-based hardware-aware model configuration search under latency, energy, and resource constraints.

\subsection{Onboard-Friendly Input Representation}
\label{subsec: Input Representation}

Our study is based on the data collected by Yoshida et al.\cite{yoshida2023smatable2}, which captures vibration signals from four synchronized piezoelectric sensors embedded in a table-mounted surface. The gestures consist of four swipe directions (Up, Down, Left, Right), sampled at 44.1 kHz and stored as 16-bit PCM waveforms. Two subsets are available: \emph{DataByPerson}, where three users (A, B, C) perform gestures on the same table, and \emph{DataByTable}, where person A interacts with three different tables (A, B, C) of varying height. Each gesture spans 2 seconds per sensor, totaling 352{,}800 samples per gesture. The original \emph{Smatable} system~\cite{yoshida2023smatable2} used STFT-based spectrograms (shape: 4096$\times$90) as input to a large 2D-CNN, offering strong accuracy but incurring high compute and memory overhead.

To eliminate costly preprocessing, we directly use raw waveform signals and apply a series of deployment-oriented optimizations.
First, each gesture record is truncated from 2 seconds to 1 second, based on the empirical observation that the majority of gesture-relevant signal energy is concentrated between 0.25 and 1.25 seconds, reducing the input to 176{,}400 samples.
Subsequently, a comprehensive empirical analysis is conducted to evaluate the trade-offs associated with various downsampling factors in terms of recognition accuracy and hardware deployability. A factor of 10 (i.e., retaining one sample every ten points) is identified as the most favorable configuration, resulting in an input shape of 4410$\times$4. This choice achieves 21$\times$ reduction in input dimensionality compared to the spectrogram-based input, significantly lowering memory and computational costs for embedded FPGA inference.
To further mitigate potential information loss introduced by downsampling, sliding-window augmentation with fractional offsets is applied. This augmentation strategy leverages previously unsampled waveform segments to expand the training dataset by around 10$\times$, thereby improving model robustness and generalization without introducing artificial distortions.

\subsection{Efficient Model Design}
\label{subsec:model_design}

Building on the lightweight waveform input, we further improve deployability through efficient NNs design for embedded FPGAs. We develop two compact 1D convolutional architectures: a 1D-CNN and a depthwise-separable variant, 1D-SepCNN. Both architectures operate on downsampled waveforms and are optimized to reduce memory footprint and arithmetic complexity under hardware constraints.
Table~\ref{tab:1d-CNN} presents an example configuration of 1D-CNN with three convolutional blocks ($\text{num}_\text{blocks}\!=\!3$). In contrast to the 2D-CNN from~\cite{yoshida2023smatable2}, which contains 369 million parameters and operates on spectrograms of shape $(4096 \times 90)$, our model directly processes downsampled waveforms of shape $(4410 \times 4)$ and contains only 296 parameters, achieving a drastic reduction in model size without significant loss in accuracy.

Each block applies a \texttt{Conv1D} layer (kernel size 3, stride 1), followed by batch normalization and ReLU. To reduce intermediate buffer size, \texttt{MaxPool1D} (kernel size 2) is applied after each block except the last, progressively reducing temporal resolution while preserving key features. The first two blocks use a narrow output channel size of 4 to minimize BRAM footprint. In deeper configurations (i.e., when $\text{num}_\text{blocks} > 3$), the number of output channels can be doubled every two blocks (e.g., 4 $\rightarrow$ 8 $\rightarrow$ 16) to gradually increase representational capacity as the temporal dimension shrinks. For example, in Table~\ref{tab:1d-CNN}, the third block increases channel width to 8 to capture higher-level gesture patterns. A global average pooling layer (\texttt{GlobalAVGPool}) aggregates features along the temporal axis, producing a compact feature vector. This vector is then passed through two \texttt{Dense} layers: the first applies ReLU activation, and the second generates logits for the four gesture classes.

\begin{table}[htbp]
\centering
\caption{Example of a 1D-CNN with three convolutional blocks. K: kernel size, C: output channels, $\rightarrow$: stride}
\label{tab:1d-CNN}
\begin{tabular}{|c|c|}
\hline
\multicolumn{2}{|c|}{Input: 4410$\times$4} \\ \hline

\multirow{4}{*}{Block1} 
& \cellcolor{blue!25}Conv1D K:3, C:4, $\rightarrow$ 1 \\ \cline{2-2}
& \cellcolor{pink!25}BatchNormalization \\ \cline{2-2}
& ReLU \\ \cline{2-2}
& \cellcolor{orange!25}MaxPool1D K:2 \\\hline

\multirow{4}{*}{Block2} 
& \cellcolor{blue!25}Conv1D K:3, C:4, $\rightarrow$ 1 \\ \cline{2-2}
& \cellcolor{pink!25}BatchNormalization \\ \cline{2-2}
& ReLU \\ \cline{2-2}
& \cellcolor{orange!25}MaxPool1D K:2\\\hline

\multirow{3}{*}{Block3} 
& \cellcolor{blue!25}Conv1D K:3, C:8, $\rightarrow$ 1 \\ \cline{2-2}
& \cellcolor{pink!25}BatchNormalization \\ \cline{2-2}
& ReLU \\ \hline

\multicolumn{2}{|c|}{\cellcolor{green!25}GlobalAVGPool C:8, $\rightarrow$ 1}   \\ \hline
\multicolumn{2}{|c|}{\cellcolor{yellow!25}Dense C:4}   \\ \hline
\multicolumn{2}{|c|}{ReLU}   \\ \hline
\multicolumn{2}{|c|}{\cellcolor{yellow!25}Dense C:4}   \\ \hline
\multicolumn{2}{|c|}{Output: [Up, Down, Left, Right]}   \\ \hline
 
\end{tabular}
\end{table}

To further reduce computational cost, we implement a variant using depthwise-separable convolution (\texttt{SepConv1D}). Each standard convolution is replaced by a \texttt{DepthConv1D} (applying a separate filter per input channel) followed by a \texttt{PointConv1D} (kernel size 1) that fuses information across channels~\cite{mo2021fpga}. 
This design substantially reduces parameter count and \emph{Floating-Point Operations} (FLOPs). For example, replacing the \texttt{Conv1D} in the first block lowers parameters from 52 to 36 and FLOPs from 84{,}576 to 49{,}336. When applied to all three blocks, the total parameter count decreases to 216 (27.0\% reduction), and FLOPs to 383{,}684 (33.1\% reduction). These savings make 1D-SepCNN especially suitable for compute-bound FPGAs, where minimizing arithmetic workload is critical to achieving real-time performance.

\subsection{Model Quantization and RTL Implementation}
\label{subsec:quantization}

Building on the efficient model architectures, we apply quantization and RTL generation to enable integer-only inference on embedded FPGAs, where all numerical components (including weights, biases, and activations) and arithmetic operations are implemented using integer arithmetic~\cite{Jacob2017}. This eliminates any floating-point (FP32) dependencies, significantly reducing hardware complexity and energy consumption~\cite{Krishnamoorthi2018}. We apply signed asymmetric quantization to both weights and activations to better utilize the available dynamic range. In contrast, bias terms and BatchNorm offsets are quantized symmetrically around zero for simplicity and hardware efficiency, as their distributions are typically centered and less sensitive to offset alignment. 

This quantization scheme is supported by the open-source \emph{ElasticAI.Creator} library\footnote{\url{https://github.com/es-ude/elastic-ai.creator/tree/add-linear-quantization}}~\cite{qian2023elasticai}, which provides end-to-end support for QAT and integer-only inference across a wide range of NN components. The library offers synthesizable VHDL templates for standard components such as Dense layer, ReLU activation function, batch normalization, and global average pooling~\cite{ling2024flowprecision, ling2024integer}.

To support convolutional architectures, we further extend the library with three additional modules: (1) \texttt{Conv1DBN}, which integrates 1D convolution with batch normalization, (2) \texttt{SepConv1DBN}, which integrates depthwise convolution \texttt{DepthConv1D} and pointwise convolution with folded batch normalization \texttt{PointConv1DBN}, and (3) \texttt{MaxPool1D}, which performs temporal downsampling. Each module has both \texttt{forward()} and \texttt{int\_forward()} modes to represent QAT and integer-only inference, respectively.

\begin{figure*}[!htb]
    \centering
    \includegraphics[width=.9\textwidth]{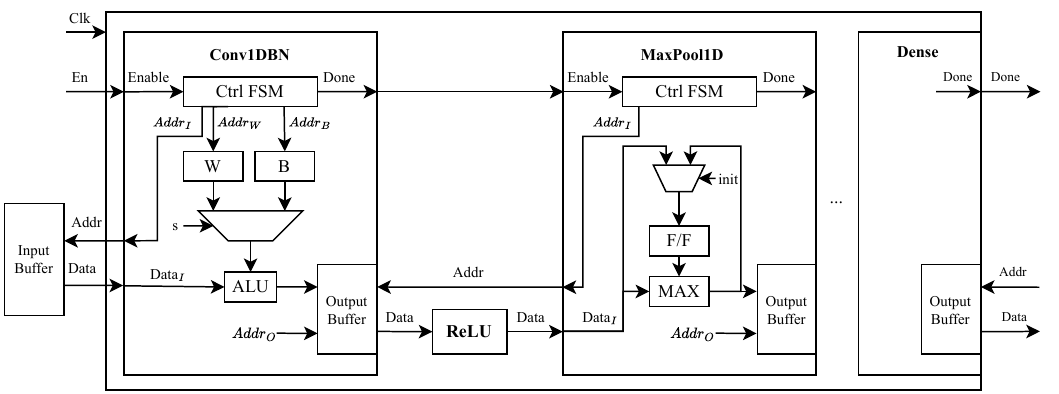}
    \caption{Schematic diagram of an 1D-CNN accelerator}
    \label{fig:hw_accelerator}
\vspace{-10pt}
\end{figure*}

All modules expose a unified \texttt{design()} interface, which the Python-based generator invokes to compile the quantized model into synthesizable RTL. This generator automatically instantiates parameterized VHDL modules for each layer, facilitating the deployment of compact 1D-CNN and 1D-SepCNN models on resource-constrained embedded FPGAs. Trained weights and biases are preloaded into on-chip memories, while quantization parameters (e.g., scale factor and zero-point) are compiled into digital logic for runtime access. All accelerator components operate in a streaming dataflow manner, communicating through a lightweight address/data protocol. Each module manages its own data readiness and handshake signals, enabling fully decoupled pipeline execution without central control. Activation functions such as \texttt{ReLU} are directly integrated into the streaming path to minimize latency and control overhead. Figure~\ref{fig:hw_accelerator} illustrates the accelerator architecture for the 1D-CNN model. The same generator flow also supports the 1D-SepCNN variant by mapping its \texttt{SepConv1DBN} layers to modular building blocks.

\begin{figure}[!htbp]
    \centering
    \begin{minipage}[t]{\linewidth}
        \centering
        \includegraphics[width=\linewidth, trim=20pt 0pt 5pt 0pt, clip]{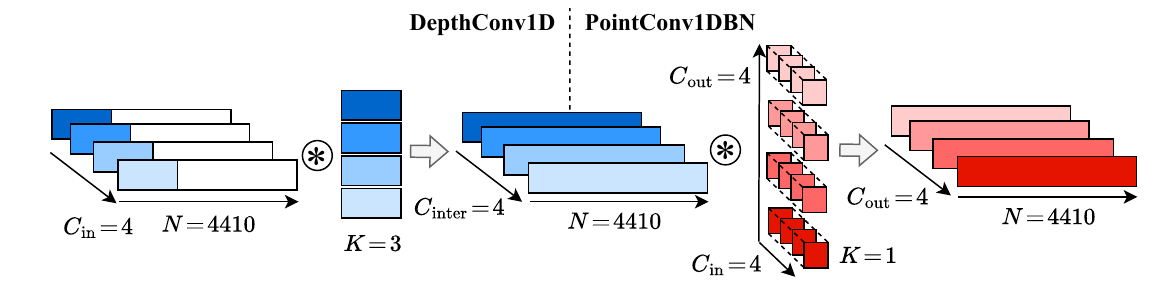}
        \caption*{(a) Without Ping-Pong}
        \vspace{15pt}
    \end{minipage}
    \begin{minipage}[t]{\linewidth}
        \centering
        \includegraphics[width=\linewidth,trim=20pt 0pt 5pt 0pt, clip]{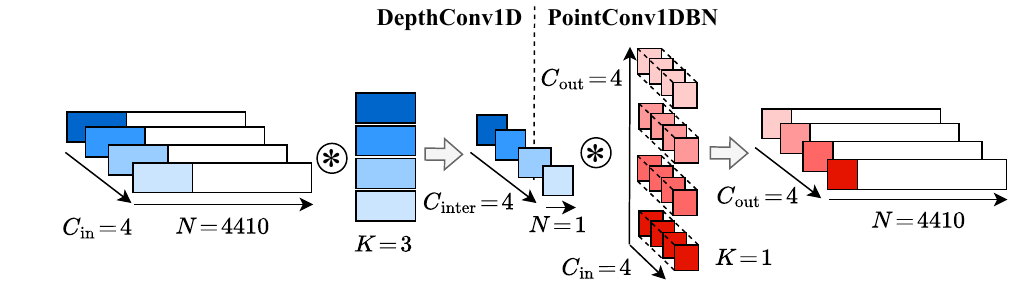}
        \caption*{(b) With Ping-Pong} 
    \end{minipage}
    \caption{\texttt{SepConv1DBN} execution without and with ping-pong buffering between the depthwise and pointwise stages}
    \label{fig:sepconv1dbn}
\vspace{-15pt}
\end{figure}

As discussed in Section~\ref{subsec:model_design}, the adoption of \texttt{SepConv1D} significantly reduces parameters and FLOPs compared to \texttt{Conv1D}, but we observed that its implementation on resource-constrained FPGAs incurs notable memory overhead. This stems from its two-stage structure: the depthwise and pointwise convolutions are executed sequentially, requiring the intermediate feature maps to be buffered between stages. In a naive implementation (see Figure~\ref{fig:sepconv1dbn}(a)), the depthwise stage must complete processing the entire input before the pointwise stage begins, requiring a buffer of size $C \times N$ (e.g., $4 \times 4410$).

To mitigate this bottleneck, we introduce a ping-pong scheduling mechanism for \texttt{SepConv1DBN} (see Figure~\ref{fig:sepconv1dbn}(b)). Instead of storing the full intermediate tensor, the depthwise stage writes one temporal slice of shape $C \times 1$ into a shared buffer and yields control to the pointwise stage, which consumes the data immediately. The two stages alternate in a synchronized manner, reducing the buffer requirement from $C \times N$ to $C \times 1$ and enabling memory-efficient streaming. For the 1D-SepCNN under 8-bit quantization (following the configuration in Table \ref{tab:1d-CNN}), BRAM utilization drops from 97.78\% to 66.67\%, and LUT usage is slightly reduced from 20.62\% to 19.74\%, with no additional DSP cost. These results highlight that memory-aware pipelining is essential for realizing the theoretical advantages of separable convolution in practice, particularly on resource-constrained FPGAs. 

\subsection{Constraint-Aware Deployment Search}
\label{subsec:con_aware}

\begin{figure*}[!htb]
    \centering
    \includegraphics[width=1\textwidth]{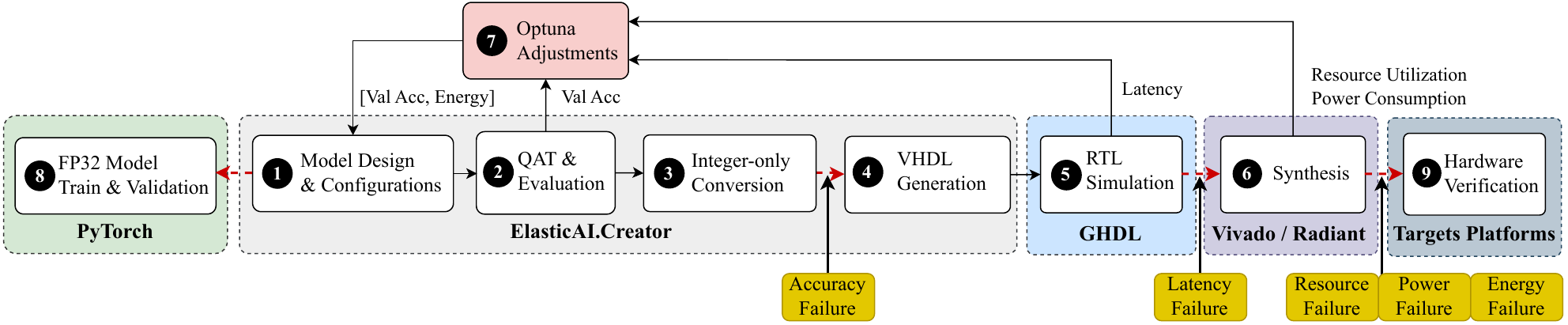}
    \caption{Overview of the deployment workflow, modified from ~\cite{ling2025deployment}}
    \label{fig:deployment_pipeline} 
\vspace{-10pt}
\end{figure*}

With the above optimizations in place, final deployment quality still hinges on selecting model configurations that balance accuracy and energy efficiency. To address this, we extend an open-source end-to-end deployment pipeline~\cite{ling2025deployment} to support the proposed 1D-CNN and 1D-SepCNN architectures. The framework builds on \emph{ElasticAI.Creator} library and integrates a hardware-aware Optuna-based search framework to form a unified deployment flow from QAT to deployable FPGA implementations. As illustrated in Figure~\ref{fig:deployment_pipeline}, it automates training, quantization, simulation, and profiling to guide the search toward viable configurations under practical constraints.

Unlike prior work~\cite{ling2025deployment} which prunes infeasible candidates based solely on resource usage, our framework applies early termination logic based on deployment-oriented criteria. Rather than completing all training and synthesis steps for every trial, constraint checks are applied progressively to prune invalid configurations as early as possible. This not only reduces redundant computation but also helps the Optuna sampler focus on promising regions of the search space. The constraint logic is fully implemented via Optuna’s trial pruning mechanism, without modifying its sampling strategy. The following constraint checks are incorporated:
\subsubsection{Accuracy Constraints}
To avoid wasting training resources on configurations unlikely to meet accuracy requirements, we define accuracy thresholds based on the minimum acceptable performance for the target task. Trials failing to meet the required accuracy during the early training phase are pruned before RTL simulation.

\subsubsection{Latency Constraint}
For real-time interaction applications, we impose a latency threshold of 100~ms, as introduced in Section~\ref{sec:introduction}. Trials exceeding this bound during post-training simulation are halted before synthesis and power analysis.

\subsubsection{Resource Constraints}
Trials resulting in non-deployable implementations due to resource overuse (e.g., LUTs, DSPs, BRAMs) are pruned before power estimation.

\subsubsection{Hardware Constraint} 
To ensure deployability under energy and power budgets, we impose a maximum power threshold of 500~mW and an energy threshold of 50~mJ per inference. Trials that exceed either bound, or for which hardware profiling fails to return valid estimates (e.g., due to synthesis failure or missing reports), are marked as infeasible and pruned accordingly.

This staged pruning methods enables efficient screening of unqualified designs. While all thresholds are manually defined in this study, they can be made adaptive in future work.

\section{Results and Evaluations}
\label{sec:results_eval}

The section starts by outlining the dataset splitting methods used in all experiments. We then reproduce the full-precision (FP32) 2D-CNN baselines from~\cite{yoshida2023smatable2} under identical settings to establish fair references. Next, our constraint-aware, hardware-aware optimization framework is applied to search for efficient and deployable configurations of the proposed 1D-CNN and 1D-SepCNN models. Finally, we evaluate how well the selected configurations generalize across datasets.

\subsection{Datasets Splitting Methods}

We adopt both the \emph{DataByPerson} and \emph{DataByTable} datasets\footnote{\url{https://zenodo.org/records/17275491}} from~\cite{yoshida2023smatable2}. Each dataset comprises nine recording sessions per target (i.e., person or table). Within each session, four swipe directions are performed. For each direction, 10 distinct recordings are collected, resulting in 40 multichannel waveform recordings per session. As mentioned in Section~\ref{subsec: Input Representation}, we apply sliding-window augmentation with fractional offsets during downsampling, which expands each session from 40 to 400 recordings. We also adopt the data splitting methods in \cite{yoshida2023smatable2}: (1) \emph{Per-Subject} (PS): Each target subject is split independently, with six sessions used for training and the remaining three for testing. (2) \emph{Leave-One-Subject-Out} (LOSO): The model is trained on two targets and evaluated on the third, simulating cross-person or cross-table generalization. (3) \emph{Add-One-Session} (AOS): One session from the held-out target is included in the training set, reflecting a practical scenario where limited target-specific data is available.

\subsection{Experiment 1: FP32 2D-CNN Baseline Reproduction}
\label{subsec: exp1}

In~\cite{yoshida2023smatable2}, Yoshida et al. revised their 2D-CNN architecture and increased the number of STFT segments from 1000 to 4096, reporting improved classification accuracy compared to their earlier work~\cite{yoshida2023smatable1}. However, only partial results were provided, specifically, test accuracies of 0.67 (LOSO) and 0.90 (AOS) on the \emph{DataByPerson} dataset, averaged over three users.
To establish a more comprehensive performance reference, we reproduce the FP32 2D-CNN model under the exact experimental settings described in their work and evaluate it across all three data splitting methods (PS, LOSO, AOS) on both the \emph{DataByPerson} and \emph{DataByTable} datasets.

\begin{table}[!htb]
\centering
\caption{Reproduced baseline test accuracies of the FP32 2D-CNN model from~\cite{yoshida2023smatable2}}
\label{tab:exp1}
\renewcommand{\arraystretch}{1.1}
\begin{tabular}{|c|cc|c|c|}
\hline
\multirow{2}{*}{\begin{tabular}[c]{@{}c@{}}\textbf{Splitting} \\ \textbf{Method}\end{tabular}} & \multicolumn{1}{c|}{\multirow{2}{*}{\begin{tabular}[c]{@{}c@{}}\textbf{Training} \\ \textbf{Data}\end{tabular}}} & \multirow{2}{*}{\begin{tabular}[c]{@{}c@{}}\textbf{Testing}\\ \textbf{Data}\end{tabular}} & \multicolumn{2}{c|}{\textbf{Dataset}}  \\ \cline{4-5} 
& \multicolumn{1}{c|}{} & & \multicolumn{1}{c|}{\textbf{ByPerson}} & \textbf{ByTable} \\ \hline

\multirow{4}{*}{PS}                                & \multicolumn{1}{c|}{A} & A & 1.000 & 1.000  \\ \cline{2-5} 
& \multicolumn{1}{c|}{B} & B & 1.000 & 0.958   \\ \cline{2-5} 
& \multicolumn{1}{c|}{C} & C & 0.983 & 0.983   \\ \cline{2-5} 
& \multicolumn{2}{c|}{avg.}  & 0.994 & 0.980  \\ \hline

\multirow{4}{*}{LOSO}                              & \multicolumn{1}{c|}{B,C} & A &  0.649 & 0.683 \\ \cline{2-5} 
& \multicolumn{1}{c|}{A,C} & B &  0.699 & 0.241 \\ \cline{2-5} 
& \multicolumn{1}{c|}{A,B} & C &  0.666 & 0.574 \\ \cline{2-5} 
& \multicolumn{2}{c|}{avg.}    &  0.671 & 0.499 \\ \hline

\multirow{4}{*}{AOS}                               & \multicolumn{1}{c|}{B,C + 1A} & A & 0.941 & 0.850 \\ \cline{2-5} 
& \multicolumn{1}{c|}{A,C + 1B} & B & 0.800 & 0.733 \\ \cline{2-5} 
& \multicolumn{1}{c|}{A,B + 1C} & C & 0.949 & 0.941 \\ \cline{2-5} 
& \multicolumn{2}{c|}{avg.}         & 0.897 & 0.841 \\ \hline
\end{tabular}
\end{table}

As shown in Table~\ref{tab:exp1}, our reproduced results on \emph{DataByPerson} achieve average test accuracies of 0.994 (PS), 0.671 (LOSO), and 0.897 (AOS), closely matching the reported values in~\cite{yoshida2023smatable2}. Minor deviations in significant digits are observed but remain within acceptable variance.
On \emph{DataByTable}, which was not evaluated in~\cite{yoshida2023smatable2}, the model performs well under PS (0.980), but shows significantly lower accuracy under LOSO (0.499) and moderate performance under AOS (0.841), indicating limited cross-table generalizability. Compared to their earlier work~\cite{yoshida2023smatable1}, these reproduced results yield mixed outcomes: occasionally improving accuracy, but often failing to generalize across different tables. These findings highlight the limitations of large, STFT-based 2D-CNN models in table-shifted scenarios, reinforcing the need for lightweight, robust alternatives that generalize better and support efficient deployment on resource-constrained platforms.

\begin{figure*}[t]
\centering
\begin{minipage}[t]{\textwidth}
    \centering
    \captionof{table}{Selected model configurations with best integer-only test accuracy on person A of DataByPerson dataset.}
    \label{tab:exp2}
    \resizebox{\textwidth}{!}{%
    \renewcommand{\arraystretch}{1.2}
\begin{tabular}{|c|c|c|c|c|c|c|c|c|c|c|c|c|c|}
\hline

\multirow{2}{*}{\begin{tabular}[c]{@{}c@{}}\textbf{Splitting} \\ \textbf{Method}\end{tabular}} & \multirow{2}{*}{\textbf{Model}} & \multicolumn{4}{c|}{\textbf{Configuration}} & \multicolumn{2}{c|}{\textbf{Test Accuracy}} & \multirow{2}{*}{\begin{tabular}[c]{@{}c@{}}\textbf{LUTs}\\ \textbf{(\%)}\end{tabular}} & \multirow{2}{*}{\begin{tabular}[c]{@{}c@{}}\textbf{BRAMs}\\ \textbf{(\%)}\end{tabular}} & \multirow{2}{*}{\begin{tabular}[c]{@{}c@{}}\textbf{DSPs}\\ \textbf{(\%)} \end{tabular}} & \multirow{2}{*}{\begin{tabular}[c]{@{}c@{}}\textbf{Latency$^{*}$}\\ \textbf{(ms)} \end{tabular}} & \multirow{2}{*}{\begin{tabular}[c]{@{}c@{}}\textbf{Power$^{**}$}\\ \textbf{(mW)}\end{tabular}} & \multirow{2}{*}{\begin{tabular}[c]{@{}c@{}}\textbf{Energy}\\ \textbf{(mJ)}\end{tabular}} \\ \cline{3-8}

&  & \textbf{ $\text{num}_\text{blocks}$} & \textbf{b} & \textbf{bs} & \textbf{lr ($\times10^{-4})$} & \multicolumn{1}{c|}{\textbf{FP32}} & \textbf{Quantized} &  & & & & & \\ \hline

\multirow{2}{*}{PS} 
& 1D-CNN    & 3 & 6 & 32 & 5.082 & 0.998 & 0.996 ($\downarrow$0.20\%) & 13.35 & 50.00 & 7.50 & 9.22  & 129  & 1.189 \\ \cline{2-14} 
& 1D-SepCNN & 3 & 8 & 48 & 5.550 & 0.952 & 0.952 ($-$0.00\%) & 19.74 & 66.67 & 11.25 & 6.83 & 163 & 1.113 \\ \hline 

\multirow{2}{*}{LOSO} 
& 1D-CNN    & 5 & 6 & 56 & 3.251 & 0.744 & 0.738 ($\downarrow$0.81\%) & 18.98 & 73.33 & 10.00  &  20.94 & 152  & 3.182\\ \cline{2-14} 
& 1D-SepCNN & 3 & 6 & 56 & 3.330 & 0.702 & 0.675 ($\downarrow$3.85\%) & 16.08 & 50.00 & 11.25 & 6.83 & 146 &  0.996 \\ \hline 

\multirow{2}{*}{AOS} 
& 1D-CNN    & 4 & 8 & 32 & 4.810 & 0.948 & 0.941 ($\downarrow$0.74\%) & 19.08 & 83.33 & 8.75 & 13.32  & 159 & 2.118  \\ \cline{2-14} 
& 1D-SepCNN & 5 & 8 & 48 & 9.967 & 0.930 & 0.909 ($\downarrow$2.26\%) & 29.82 & 96.67 & 16.25 & 11.16 & 199 & 2.221 \\ \hline 

\multicolumn{14}{l}{$\text{num}_\text{block}$ = the number of convolutional blocks, $b$ = quantization bitwidth, $bs$ = batch size, $lr$ = learning rate.} \\

\multicolumn{14}{l}{$^{*}$ Latency measured on actual hardware deviates by 1.95\% from simulation results.}\\

\multicolumn{14}{l}{$^{**}$ Power estimated from Vivado synthesis reports was validated on the actual FPGA hardware at 28.0\textdegree C, showing a deviation within 5.6\%.} \\
\end{tabular}
}
\end{minipage}
\begin{minipage}[t]{\textwidth}
    \vspace{10pt}
    \centering
    \begin{subfigure}[b]{0.32\textwidth}
        \includegraphics[width=\linewidth]{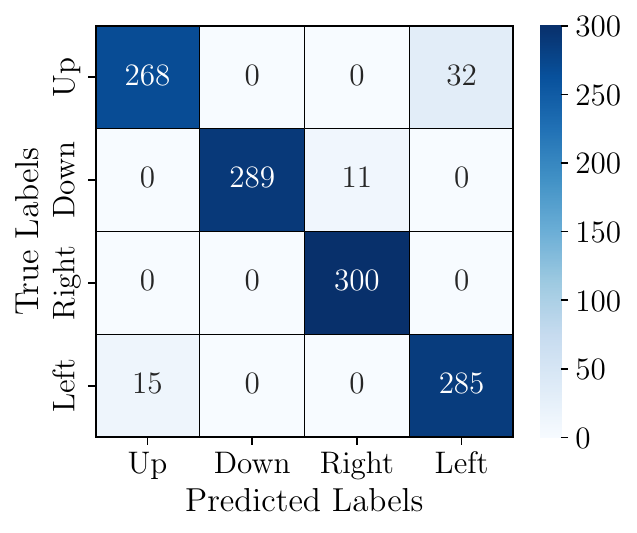}
        \caption{PS}
        \label{fig:cm_ps}
    \end{subfigure}
    \hfill
    \begin{subfigure}[b]{0.32\textwidth}
        \includegraphics[width=\linewidth]{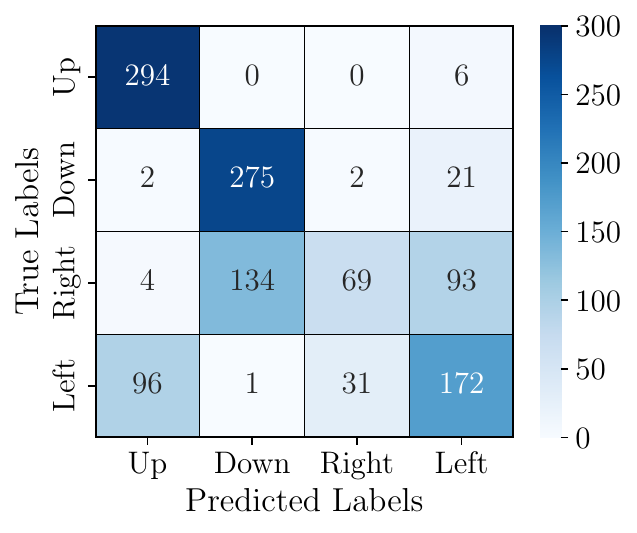}
        \caption{LOSO}
        \label{fig:cm_loso}
    \end{subfigure}
    \hfill
    \begin{subfigure}[b]{0.32\textwidth}
        \includegraphics[width=\linewidth]{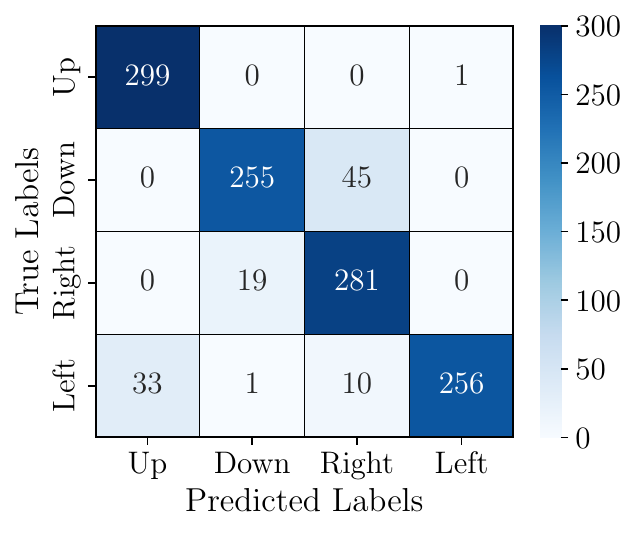}
        \caption{AOS}
        \label{fig:cm_aos}
    \end{subfigure}
    \captionof{figure}{Confusion matrices of 1D-SepCNN models from Table~\ref{tab:exp2} across data splitting methods.}
    \label{fig:1dsepcnn_confusion_matrix}
\end{minipage}
\vspace{-10pt}
\end{figure*}

\subsection{Experiment 2: Hardware-Aware Optimization}
\label{subsec:exp2}

To evaluate the effectiveness of our proposed optimizations in Section~\ref{sec:proposed_method}, we conduct a hardware-aware model configuration search using Optuna with constraint-aware bi-objective on person A of \emph{DataByPerson} dataset. This experiment aims to identify energy-efficient and deployable configurations for the 1D-CNN and 1D-SepCNN models under all three data splitting methods. The search process jointly considers software-level factors (e.g., architecture, training settings, and quantization), accuracy metrics, and hardware-level metrics (e.g., latency, power, and resource usage), operating on augmented, downsampled raw waveforms.

We perform 100 Optuna trials per combination of model type and data split method, using the \texttt{NSGAII} sampler to jointly maximize validation accuracy and minimize energy consumption per inference. Validation accuracy is obtained during QAT and evaluation (\ding{183} in Figure~\ref{fig:deployment_pipeline}). Energy consumption is computed as $E \!=\! T \times P$ in millijoules (mJ), where inference latency $T$ in milliseconds (ms) is measured via RTL simulation using GHDL (\ding{186} in Figure~\ref{fig:deployment_pipeline}), and post-synthesis power $P$ in milliwatts (mW) is estimated from Vivado 2019.2 reports following logic synthesis and static timing analysis (\ding{187} in Figure~\ref{fig:deployment_pipeline}). Only configurations satisfying all constraints (described in Section~\ref{subsec:con_aware}) are retained for further Pareto front analysis, where non-dominated configurations are extracted from the Optuna study to visualize trade-offs among accuracy and energy consumption.
The search space of Optuna study includes: (1) Quantization bitwidth: $b \in \{4, 6, 8\}$; (2) Batch size: $bs \in \{16, 24, \dots, 64\}$; (3) Learning rate: $lr \in [10^{-5}, 10^{-3}]$ (log-uniform); (4) Number of convolutional blocks: $\text{num}\text{block} \in \{1,2, \dots, 5\}$. Details of the accuracy thresholds for all data splitting settings and quantization bitwidths are provided in Table \ref{tab:acc-thresholds} in the Appendix.

Each trial is trained for up to 100 epochs using Adam optimizer with early stopping (patience \!=\! 10). Models are implemented in Python 3.11 and trained on an NVIDIA RTX 2080 SUPER (CUDA 11.0). FPGA synthesis and hardware validation (\ding{187} and \ding{190} in Figure~\ref{fig:deployment_pipeline}) are conducted on the AMD Spartan-7 family. We evaluated multiple device variants with our deployment pipeline. We finally selected the XC7S25 as the target device, as it offers the best trade-off between deployability and energy efficiency. This FPGA provides 14,600 LUTs, 1,620 Kbits of BRAM (45 blocks), and 80 DSP slices. All models are synthesized to operate at a fixed clock frequency of 100~MHz. Hardware measurements confirmed the reliability of our simulation-based estimates, showing only 1.95\% deviation in latency and 5.6\% in power.

Table~\ref{tab:exp2} summarizes model configurations, selected based on the highest test accuracy achieved during integer-only inference from Pareto fonts. Notably, the FP32 results are obtained from purely PyTorch-based models trained and validated with identical configurations, serving as baselines (\ding{189} in Figure~\ref{fig:deployment_pipeline}).
Across all settings, selected models use either 6-bit or 8-bit quantization, as 4-bit configurations consistently fail to meet the accuracy constraint. Compared to their FP32 counterparts, the quantized models exhibit only minor accuracy degradation (up to 3.85\%). Specifically, accuracy loss is minimal under PS and becomes more pronounced under LOSO.
Compared to the FP32 2D-CNN baseline in Experiment~1, both quantized 1D-CNN and 1D-SepCNN models deliver comparable or even higher accuracy across all data splitting methods, despite their limited model complexity for smaller energy consumption. For example, under the PS setting, the selected 6-bit quantized 1D-CNN achieves a test accuracy of 0.996 using only 264 parameters (after folding all Batch Normalization layers), compared to 369 million parameters in the 2D-CNN with an accuracy of 0.994. The chosen 8-bit quantized 1D-SepCNN further reduces the parameters to 184, but incurs a slightly larger accuracy drop (0.952 vs. 0.994).

Comparing 1D-SepCNN and 1D-CNN, we found that 1D-SepCNN configurations tend to be more resource-intensive than 1D-CNN in PS and AOS settings,  e.g., 29.82\% vs. 19.08\% LUT usage in AOS. However, they consistently achieve faster inference, owing to the computational efficiency of the depthwise-separable design. This yields comparable energy efficiency despite the increased power consumption of 1D-SepCNN configurations. In contrast, LOSO reveals a different trade-off: the selected 1D-SepCNN is architecturally simpler (fewer blocks, lower resource usage), resulting in drastically reduced latency (6.83 ms vs. 20.94 ms) and energy (0.996 mJ vs. 3.182 mJ). However, it suffers the largest accuracy drop (–3.85\%), likely due to overfitting in deeper configurations given limited subject diversity, as indicated by learning curve analysis. As shown in Figure~\ref{fig:1dsepcnn_confusion_matrix}(b), in LOSO setting, the model confuses adjacent gestures like ``Right'' and ``Left'' (231 and 128 misclassifications, respectively), whereas PS shows clean class separation, as displayed in Figure~\ref{fig:1dsepcnn_confusion_matrix}(a). Moreover, Figure~\ref{fig:1dsepcnn_confusion_matrix}(c) demonstrates that AOS exhibits moderate confusion, primarily between ``Down'' and ``Right'', but remains within acceptable deployment bounds.

In addition, prior work~\cite{yoshida2023smatable2} reports that training their 2D-CNN model on six sessions of vibration data took approximately 7,611 seconds on a CPU. In contrast, each trial in our hardware-aware search (up to 100 epochs) completes in an average of 479.89 seconds for 1D-CNN and 472.99 seconds for 1D-SepCNN on an RTX 2080 SUPER GPU, demonstrating at least an order-of-magnitude improvement in training efficiency per deployable configuration. 
Regarding inference time, Yoshida reports a total latency of 365 ms, comprising 15 ms for STFT-based feature extraction and 350 ms for 2D-CNN inference. Our FPGA-deployed models achieve significantly lower inference latency. For instance, as low as 6.83 ms, representing a speedup of more than 53$\times$ over Yoshida’s. This drastic improvement in latency is achieved alongside substantial reductions in model size and power consumption, highlighting the real-time responsiveness and deployability of our embedded solution.

\subsection{Experiment 3: Generalization Evaluation}

To evaluate the generalizability of the hardware-aware model configurations from Experiment~2, we apply the selected model configurations from person A on the \emph{DataByPerson} dataset (Table~\ref{tab:exp2}) to other subjects and both datasets. 

Table~\ref{tab:exp3} summarizes the best test accuracies of quantized 1D-CNN and 1D-SepCNN models under all three data splitting methods, each trained by 50 sessions. Overall, both architectures demonstrate strong generalization. 
In PS setting, 1D-CNN achieves average accuracies of 0.970 (DataByPerson) and 0.971 (DataByTable), while 1D-SepCNN reaches 0.949 and 0.964, respectively. These results demonstrate that configurations optimized on a single subject remain effective across other users and tables. Compared to the FP32 2D-CNN baseline in Experiment~1 (see Table \ref{tab:exp1}), both quantized models offer comparable accuracy across broader test cases on embedded FPGA, while reducing model size by over five orders of magnitude.

\begin{table}[!htb]
\centering
\caption{Test accuracies of the quantized 1D-CNN and 1D-SepCNN models, configured according to Table~\ref{tab:exp2}, evaluated under three data splitting methods on two datasets.}
\label{tab:exp3}
\renewcommand{\arraystretch}{1.1}
\begin{adjustbox}{max width=\columnwidth}
\begin{tabular}{|c|c|cc|c|c|}
\hline

\multirow{2}{*}{\textbf{Model}} & \multirow{2}{*}{\begin{tabular}[c]{@{}c@{}}\textbf{Splitting} \\ \textbf{Method}\end{tabular}} & \multicolumn{1}{c|}{\multirow{2}{*}{\begin{tabular}[c]{@{}c@{}}\textbf{Training} \\ \textbf{Data}\end{tabular}}} & \multirow{2}{*}{\begin{tabular}[c]{@{}c@{}}\textbf{Testing}\\ \textbf{Data}\end{tabular}} & \multicolumn{2}{c|}{\textbf{Dataset}} \\ \cline{5-6} 
 & & \multicolumn{1}{c|}{} &  & \multicolumn{1}{c|}{\textbf{ByPerson}} & \textbf{ByTable} \\ \hline
                                                                 
\multirow{12}{*}{1D-CNN}
&\multirow{4}{*}{PS} 
  & \multicolumn{1}{c|}{A} & A & 0.996 &  0.994   \\\cline{3-6} 
& & \multicolumn{1}{c|}{B} & B & 0.952 &  0.921   \\\cline{3-6} 
& & \multicolumn{1}{c|}{C} & C & 0.961 &  0.999   \\\cline{3-6} 
& & \multicolumn{2}{c|}{avg.}  & 0.970 &  0.971   \\\cline{2-6} 

& \multirow{4}{*}{LOSO} 
  & \multicolumn{1}{c|}{B,C} & A & 0.738 &  0.796   \\\cline{3-6} 
& & \multicolumn{1}{c|}{A,C} & B & 0.816 &  0.642   \\\cline{3-6} 
& & \multicolumn{1}{c|}{A,B} & C & 0.882 &  0.711   \\\cline{3-6} 
& & \multicolumn{2}{c|}{avg.}    & 0.812 &  0.716   \\\cline{2-6} 

& \multirow{4}{*}{AOS} 
  & \multicolumn{1}{c|}{B,C + 1A} & A & 0.941 & 0.991  \\\cline{3-6} 
& & \multicolumn{1}{c|}{A,C + 1B} & B & 0.872 & 0.836  \\\cline{3-6} 
& & \multicolumn{1}{c|}{A,B + 1C} & C & 0.978 & 0.998  \\\cline{3-6} 
& & \multicolumn{2}{c|}{avg.}         & 0.930 & 0.942  \\\hline

\multirow{12}{*}{1D-SepCNN}
&\multirow{4}{*}{PS} 
  & \multicolumn{1}{c|}{A} & A & 0.952 &  0.989   \\\cline{3-6} 
& & \multicolumn{1}{c|}{B} & B & 0.943 &  0.905   \\\cline{3-6} 
& & \multicolumn{1}{c|}{C} & C & 0.952 &  0.998   \\\cline{3-6} 
& & \multicolumn{2}{c|}{avg.}  & 0.949 &  0.964   \\\cline{2-6} 

& \multirow{4}{*}{LOSO} 
  & \multicolumn{1}{c|}{B,C} & A & 0.675 &  0.714   \\\cline{3-6} 
& & \multicolumn{1}{c|}{A,C} & B & 0.690 &  0.540   \\\cline{3-6} 
& & \multicolumn{1}{c|}{A,B} & C & 0.695 &  0.591   \\\cline{3-6} 
& & \multicolumn{2}{c|}{avg.}    & 0.687 &  0.615   \\\cline{2-6} 

& \multirow{4}{*}{AOS} 
  & \multicolumn{1}{c|}{B,C + 1A} & A & 0.909 & 0.973  \\\cline{3-6} 
& & \multicolumn{1}{c|}{A,C + 1B} & B & 0.888 & 0.815  \\\cline{3-6} 
& & \multicolumn{1}{c|}{A,B + 1C} & C & 0.978 & 0.995  \\\cline{3-6} 
& & \multicolumn{2}{c|}{avg.}         & 0.925 & 0.928  \\\hline

\end{tabular}
\end{adjustbox}
\vspace{-10pt}
\end{table}

In the more challenging LOSO setting, 1D-CNN maintains strong performance with 0.812 (DataByPerson) and 0.716 (DataByTable), significantly outperforming 1D-SepCNN (0.687 and 0.615, respectively). This mirrors earlier findings from Experiment~2 and reinforces the sensitivity of depthwise-separable architectures to cross-subject variation. Nonetheless, 1D-SepCNN exceeds the reported accuracy of the original 2D-CNN baseline in Experiment~1 in most cases, especially in \emph{DataByTable} dataset.
Under the AOS setting, both models exhibit strong and balanced generalization. 1D-CNN achieves 0.930 (DataByPerson) and 0.942 (DataByTable), while 1D-SepCNN attains 0.925 and 0.928, respectively. The narrowing gap between the two models suggests that 1D-SepCNN benefits significantly from even a small amount of subject-specific data. These results support our earlier observation: while 1D-SepCNN may overfit under limited training diversity, it becomes highly competitive when exposed to broader variation in users and tables.

\section{Conclusion and Future Work}
\label{sec:conclusion_future_work}

This work presents a fully FPGA-deployable vibration-based gesture recognition solution for smart surfaces, targeting low-latency, energy-efficient interaction without reliance on costly preprocessing or high-performance CPUs with high energy usage. To this end, we propose a series of end-to-end software-hardware co-optimizations.
We firstly eliminate complex spectral transformations by operating directly on downsampled raw vibration signals, reducing input size by 21$\times$ while preserving gesture-discriminative patterns. We then design two lightweight 1D convolutional architectures (1D-CNN and 1D-SepCNN) that dramatically reduce parameter counts (down to 216) compared to the 2D-CNN baseline (369 million), enabling compact yet expressive inference on resource-limited hardware. We extend the \emph{ElasticAI.Creator} library to support these models for integer-only quantization and template-based RTL design generation. In addition, we integrate a hardware-aware Optuna framework to enable constraint-driven selection of model configurations that satisfy bi-objective deployment goals.
Experimental results confirm that all selected configurations run under 21 ms and consume less than 3.2 mJ on the AMD Spartan-7 XC7S25 FPGA. The most efficient setting (1D-SepCNN, PS split) completes inference in under 7 ms with energy consumption smaller than 1.2 mJ, offering over 53$\times$ speedup compared to the CPU baseline~\cite{yoshida2023smatable2}. While 1D-CNN generalizes better across users and tables, 1D-SepCNN delivers superior latency and energy efficiency, highlighting a trade-off between robustness and deployment cost. Together, these results validate the feasibility of real-time gesture recognition on low-power FPGA platforms.

For future work, we aim to extend the system to support more gestures, such as continuous, compound, and multi-finger gestures. We also plan to explore deployment across a wider variety of furniture with diverse materials and geometries. Furthermore, we will investigate hybrid model architectures that combine convolutional, recurrent, and attention mechanisms to better adapt to diverse gesture patterns and hardware constraints, aiming for improved robustness, flexibility, and energy efficiency in real-world deployments. In addition, we intend to transition from offline gesture recognition to real-time online inference, with the goal of enabling fully integrated smart furniture systems that operate robustly in everyday home environments. 

\section*{Acknowledgments} 
 
The authors gratefully acknowledge the financial support provided by JSPS KAKENHI JP25K03107 and the Federal Ministry for Economic Affairs and Climate Action of Germany for the RIWWER project (01MD22007C).

\appendix
\begin{table}[h]
\centering
\caption{Accuracy thresholds per data splitting setting and quantization bitwidth.}
\label{tab:acc-thresholds}
\begin{tabular}{|c|c|c|c|}
\hline
\multirow{2}{*}{\textbf{Bitwidth}} & \multicolumn{3}{c|}{\textbf{Data Splitting Setting}} \\ \cline{2-4}
& \textbf{PS} & \textbf{LOSO} & \textbf{AOS} \\ \hline
8-bit  & 0.80 & 0.60 & 0.75 \\ \hline
6-bit  & 0.75 & 0.55 & 0.70 \\ \hline
4-bit  & 0.70 & 0.50 & 0.65 \\ \hline
\end{tabular}
\end{table}

\bibliographystyle{IEEEtran}
\bibliography{reference}
\end{document}